%% file: main-mlsp.tex
\documentclass{article}
\usepackage{amsmath,amssymb}
\usepackage{graphicx}
\usepackage{mlspconf}
\usepackage{booktabs}
\usepackage{xcolor}
\usepackage{tikz}
\usepackage[hidelinks]{hyperref}
\usepackage{url}

\providecommand{\citep}[1]{\cite{#1}}
\providecommand{\citet}[1]{\cite{#1}}

\providecommand{\Description}[1]{}

\makeatletter
\renewcommand{\thebibliography}[1]{\section{References}\list
 {[\arabic{enumi}]}{\settowidth\labelwidth{[#1]}\leftmargin\labelwidth
 \advance\leftmargin\labelsep
 \usecounter{enumi}%
 \setlength{\itemsep}{0pt plus 0pt minus 0pt}%
 \setlength{\parsep}{0pt plus 0pt minus 0pt}}
 \def\newblock{\hskip .11em plus .33em minus .07em}
 \sloppy\clubpenalty4000\widowpenalty4000
 \sfcode`\.=1000\relax}
\makeatother

\input{content/macros}

\title{Architecture-Induced Recoverability Bias in
  Differentiable Symbolic Regression}

\name{%
    Chakshu Gupta$^{\star}$%
    \qquad Theodore J. LaGrow$^{\star\dagger}$%
}
\address{%
    $^{\star}$ College of Computing, Georgia Institute of Technology, Atlanta, GA, USA \\%
    $^{\dagger}$ College of Lifetime Learning, Georgia Institute of Technology, Atlanta, GA, USA%
}

\begin{document}
\ninept
\maketitle

\begin{abstract}
\input{content/abstract}
\end{abstract}

\begin{keywords}
Differentiable symbolic regression, expression recovery, architecture-induced bias, gradient-based symbolic regression, nonlinear system identification
\end{keywords}

\input{content/introduction}
\input{content/background}
\input{content/methodology}
\input{content/results}
\input{content/discussion}
\input{content/conclusion}

\bibliographystyle{IEEEbib}
\bibliography{../bib/references}

\end{document}

%% file: content/macros.tex
\newcommand{\eml}{\mathrm{eml}}
\newcommand{\sml}{\mathrm{sml}}
\newcommand{\rml}{\mathrm{rml}}

\makeatletter
\renewcommand\section{\@startsection{section}{1}{\z@}%
  {-2.8ex \@plus -0.8ex \@minus -.2ex}%
  {1.8ex \@plus.2ex}%
  {\normalfont\Large\bfseries}}
\renewcommand\subsection{\@startsection{subsection}{2}{\z@}%
  {-2.4ex \@plus -0.7ex \@minus -.2ex}%
  {1.1ex \@plus.2ex}%
  {\normalfont\large\bfseries}}
\renewcommand\subsubsection{\@startsection{subsubsection}{3}{\z@}%
  {-2.1ex \@plus -0.6ex \@minus -.2ex}%
  {0.9ex \@plus.2ex}%
  {\normalfont\normalsize\bfseries}}
\makeatother

%% file: content/abstract.tex
Symbolic regression aims to recover closed-form expressions
from numerical data, but in differentiable symbolic regression
the recovered expression depends not only on the grammar but
also on the fixed architecture through which variables are
routed during training.  This is relevant to signal-processing
settings in which closed-form models and interpretable nonlinear
structure are useful.  This architecture-specific effect has
rarely been isolated directly, because existing comparisons often
vary architecture together with operator family, grammar, or
search procedure.
Three depth-3 architectures are compared across twenty-four
operator--shape--leaf combinations, holding operator family,
grammar, and training protocol fixed as far as possible while
varying the variable-routing architecture.  Recovery changes from
$0/64$ to $64/64$ trials on the same target under an
architecture-plus-native-training-protocol comparison.  The best
architecture on one target is the worst on another, and trees
with two equal-depth subtrees fail in every configuration tested
($0/3{,}776$).  As a proof-of-concept mitigation, a small
architecture set is trained and the hardened expression with the
lowest held-out RMSE is selected.  On the jointly-run subset,
this improves recovery from $34.4\%$ for the only architecture
present in all three configurations to $50.1\%$.  On a
Shockley diode target, the validation selector recovers cases
missed by that baseline architecture, which by itself recovers
$0/32$ seeds.  Since the jointly-run subset contains only three
configurations, the selector result is evidence that
validation-based architecture selection is promising, not a
complete benchmark.  These results support treating architecture
as a measurable design variable that should be reported,
stress-tested, and selected using held-out validation rather
than fixed a priori.

%% file: content/introduction.tex
\section{Introduction}
\label{sec:intro}

Closed-form models of nonlinear signal-processing systems are
useful for interpretability, sensitivity analysis, and downstream
design.  Symbolic regression (SR) recovers such models from
data~\citep{Cranmer2023pysr, LaCava2021srbench}.
Differentiable SR over a fixed tree of operator
nodes~\citep{Sahoo2018eql, Dong2024eeql, Odrzywolek2026eml} is
especially relevant when the learned expression is intended to
be inspected, simplified, or reused in downstream
signal-processing models.  The tree's structure encodes inductive
biases the practitioner controls before training.  The tree's
shape and where input variables enter the tree together form its
architecture, fixed before any data is seen.  How much that
fixed choice constrains gradient-based recovery, when operator
and grammar are held constant, remains difficult to isolate
because existing comparisons usually vary architecture together
with at least one other factor.

SRBench~\citep{LaCava2021srbench} compares 14 methods on 252
problems whose operator sets and search algorithms vary across
methods, confounding architecture with grammar and search.  The
Equation Learner (EQL)~\citep{Sahoo2018eql} and its extension
eEQL~\citep{Dong2024eeql} each evaluate a single fixed
architecture, leaving the effect of architecture choice
unmeasured.  MetaSymNet~\citep{Li2025metasymnet} and
DySymNet~\citep{Li2024dysymnet} adapt architecture during search,
quantifying end-to-end recovery rather than the contribution of
any one fixed architecture.  Methods outside the fixed-tree
gradient setting~\citep{Petersen2021dsr, Cranmer2023pysr,
Brunton2016sindy} operate on a different design space and are
not directly comparable.

Isolating variable-routing architecture requires a controlled
operator family in which the tested targets are representable
within the chosen depth limit.  The Exp-Minus-Log (EML) operator~\citep{Odrzywolek2026eml}
provides such a controlled setting for the targets studied here.  Three architectures, each a different
variable-routing pattern over a depth-3 EML tree, are compared
on a common set of targets, and two further operators, one
sharing EML's gradient asymmetry and one reversing that
asymmetry, test whether the findings depend on EML specifically
or on the gradient-asymmetry pattern exhibited by the operator.
This paper makes three contributions.  First, we formalize
recoverability as distinct from representability in fixed-tree
differentiable SR.  Second, we introduce a controlled comparison
in which variable-routing architecture is varied while operator
and grammar are held fixed.  Third, we show, as a proof of
concept, that held-out validation over a small architecture set
can mitigate some recovery failures, including on a Shockley
diode target outside the synthetic target family.

%% file: content/background.tex
\section{Background}
\label{sec:background}

In the differentiable symbolic regression variant studied here, a
fixed binary tree of operator nodes is parameterized so that
gradient descent learns the leaf weights and a soft selection of
the input that each internal node receives.  An architecture
specifies the tree's depth, branching pattern, and the routing of
input variables through the tree.

The EML operator~\citep{Odrzywolek2026eml} is the binary
expression $\eml(a,b) = e^a - \ln b$.  Setting one argument to a
constant exposes one half: $\eml(x,1) = e^x$ and
$\eml(0,x) = 1 - \ln x$.  The two halves can also be exercised
at once, as in $\eml(x,x) = e^x - \ln x$ or
$\eml(\eml(x,1),\,x) = e^{e^x} - \ln x$.  With the constant~$1$
available at leaves, EML has been proposed as a functionally
complete operator for elementary
expressions~\citep{Odrzywolek2026eml}.  In this paper, however,
the relevant question is narrower, whether the tested depth-3
targets are representable and recoverable under training and
hardening.  Although the cited operator has broad expressivity
in principle, the experiments here are depth-limited.  The
relevant question is therefore whether a target lies inside the
hardened depth-3 architecture and whether training can recover
that target.

The two arguments of $\eml$ have asymmetric partial derivatives:
$\partial\eml/\partial a = e^a$ grows exponentially with $a$,
while $\partial\eml/\partial b = -1/b$ shrinks toward zero as
$b$ grows.  By the chain rule, the gradient of the training MSE
at any parameter is the product of local derivatives along the
path from that parameter to the root.  In the observed training
regime, many internal outputs exceed one.  In that regime, a
path through more left-side ($a$) inputs accumulates
amplification, while a path through more right-side ($b$) inputs
accumulates attenuation.  Routing a variable through more
left-side inputs is therefore expected to increase its gradient
magnitude relative to right-side routing.

A binary tree of depth~$n$ has $2^n - 1$ internal nodes and
$2^n$ leaves.  Depth~3 gives seven internal nodes and eight leaves,
the template used throughout this study.  Each internal node is
given a small set of candidate inputs, and training selects one of
them, so different selections collapse the same template into
different \emph{active} subtrees.  Many active subtrees are
admissible at depth~3; this study focuses on five archetypal
shapes that span the chain and balanced extremes
(Fig.~\ref{fig:topologies}).  Four are chains: at each internal
level above the deepest, one branch continues into a subtree
while the other reduces to the constant~1; the deepest active
node takes the variable leaves $x$ and $y$.  Chain shapes are labeled by which $\eml$
argument carries the active branch at the root and at depth~1,
left ($a$) or right ($b$), giving LR, RL, RR, and LL.

The fifth is balanced.  Both root children are $\eml$ nodes that
take leaves directly, so its active subtree occupies only the top
two levels of the depth-3 template.  The LL chain produces triply iterated
exponentiation ($e^{e^{e^x}}$ at unit-leaf inputs), which
overflows on the training domain and is excluded from experiments.

All three architectures share the depth-3 template and differ
only in how each internal node selects its inputs
(Table~\ref{tab:archs}); each contains every target in this paper
inside its representable set.  Eq.\,6 is taken directly from
Eq.~(6) of~\citet{Odrzywolek2026eml}: a softmax at every internal
node selects among $\{1, x, f_{\text{child}}\}$, where
$f_{\text{child}}$ at each input position denotes the output of
the $\eml$ subtree rooted at that position, so the variable~$x$
can enter at every level but $y$ enters only at the leaves.  V16
is variant~16 of the EML reference implementation released with
the cited paper~\citep{Odrzywolek2026eml}: a
sigmoid at every internal node selects between
$\{1, f_{\text{child}}\}$, and $x$ and $y$ enter only at the
leaves through a three-way softmax over $\{x, y, 1\}$, treating
the two variables symmetrically.  These two architectures differ
in two ways at once.  They have different selector arity (softmax over three inputs
versus sigmoid over two) and different variable-routing asymmetry
(favoring~$x$ versus treating $x,y$ alike).  Hybrid localizes
this distinction to the root.  Hybrid uses V16's sigmoid at
every internal node other than the root, where a four-way
softmax additionally admits $x$ and $y$, granting both variables
direct root access without otherwise altering the V16 selector
pattern.

\begin{figure}[t]
\centering
\resizebox{\columnwidth}{!}{%
\begin{tikzpicture}[
  level distance=5mm, sibling distance=4mm,
  every node/.style={circle, draw, minimum size=3mm, inner sep=0pt, font=\scriptsize},
  active/.style={fill=black!20},
  leaf/.style={rectangle, draw, minimum size=2.5mm, font=\scriptsize},
  label/.style={draw=none, fill=none, font=\scriptsize\bfseries, minimum size=0},
]
\begin{scope}[xshift=0cm]
\node[label, above] at (0,0.3) {LR};
\node[active] (r) at (0,0) {$f$}
  child { node[active] {$f$}
    child { node[leaf] {1} }
    child { node[active] {$f$}
      child { node[leaf] {$x$} }
      child { node[leaf] {$y$} }
    }
  }
  child { node[leaf] {1} };
\end{scope}
\begin{scope}[xshift=1.7cm]
\node[label, above] at (0,0.3) {RL};
\node[active] (r) at (0,0) {$f$}
  child { node[leaf] {1} }
  child { node[active] {$f$}
    child { node[active] {$f$}
      child { node[leaf] {$x$} }
      child { node[leaf] {$y$} }
    }
    child { node[leaf] {1} }
  };
\end{scope}
\begin{scope}[xshift=3.3cm]
\node[label, above] at (0,0.3) {RR};
\node[active] (r) at (0,0) {$f$}
  child { node[leaf] {1} }
  child { node[active] {$f$}
    child { node[leaf] {1} }
    child { node[active] {$f$}
      child { node[leaf] {$x$} }
      child { node[leaf] {$y$} }
    }
  };
\end{scope}
\begin{scope}[xshift=5.1cm]
\node[label, above] at (0,0.3) {LL};
\node[active] (r) at (0,0) {$f$}
  child { node[active] {$f$}
    child { node[active] {$f$}
      child { node[leaf] {$x$} }
      child { node[leaf] {$y$} }
    }
    child { node[leaf] {1} }
  }
  child { node[leaf] {1} };
\end{scope}
\begin{scope}[xshift=6.8cm,
  level 1/.style={sibling distance=7mm},
  level 2/.style={sibling distance=3.5mm}]
\node[label, above] at (0,0.3) {Bal};
\node[active] (r) at (0,0) {$f$}
  child { node[active] {$f$}
    child { node[leaf] {$x$} }
    child { node[leaf] {1} }
  }
  child { node[active] {$f$}
    child { node[leaf] {$y$} }
    child { node[leaf] {1} }
  };
\end{scope}
\end{tikzpicture}}
\Description{Five depth-3 tree topologies: LR, RL, RR, LL (chains) and Balanced, showing which nodes carry active computation.  Internal nodes labeled $f$ to indicate the topology applies to any binary operator.}
\caption{Depth-3 active-subtree shapes used to generate target
expressions.  The labels LR, RL, RR, and LL describe left/right
routing of the active chain through the binary operator $f$; Bal
denotes the balanced two-branch topology.}
\label{fig:topologies}
\end{figure}

The targets used throughout are the closed forms produced by
substituting the variable leaves $\{x, y\}$ into each shape in
Fig.~\ref{fig:topologies}.  Two assignments per shape exist
($(x,y)$ and $(y,x)$), and the labels in the result tables abbreviate
them.  \emph{Paper}\,$(y,x)$ is the LR target
$\mathrm{eml}(\mathrm{eml}(1, \mathrm{eml}(y,x)), 1)$ used as the
single-leaf example in~\citep{Odrzywolek2026eml}, T1\,$(x,y)$ is the
RL target $\mathrm{eml}(1, \mathrm{eml}(\mathrm{eml}(x,y), 1))$,
T4\,$(x,y)$ is the RR target
$\mathrm{eml}(1, \mathrm{eml}(1, \mathrm{eml}(x,y)))$, and T8\,$(x,y)$
is the LR target with leaves swapped.  The yx variants apply the
$x \leftrightarrow y$ relabeling to the same shape.

\subsection{Recoverability framework}
\label{sec:framework}

Let $A$ denote an architecture with trainable parameters
$\Theta_A$.  These parameters are the leaf weights together with
the selector parameters at every softmax and sigmoid internal
node.  Training starts from an initial value
$\theta_0 \in \Theta_A$ drawn at random and produces a final
value $\theta_T$ by minimizing the training loss.  Each selector
is then hardened by the one-hot snap, in which every softmax and sigmoid is
replaced by the $\arg\max$ of its inputs, and branches no longer
selected are removed, leaving a concrete expression
$H(A,\theta_T)$ in closed form.

The representable set
\begin{equation}
  E(A) = \{H(A, \theta) : \theta \in \Theta_A\}
\end{equation}
collects every expression that hardening can produce.
The recoverability of a target $f^\star$ under $A$ is
\begin{equation}
  R(A, f^\star) \;=\; \Pr_{\theta_0,\,\text{train}}\bigl[\,H(A,\theta_T) \equiv f^\star\,\bigr],
  \label{eq:recover}
\end{equation}
the probability, over random initialization and the training
procedure, that the hardened expression matches $f^\star$ both
structurally and numerically.  Both
checks are necessary.  Two trees of the same shape can carry
different leaf weights and represent different functions, while
numerically close expressions need not recover the intended
symbolic structure.  Structural match compares the post-snap
tree against $f^\star$ using a canonical signature.  The
signature accounts for operator commutativity at symmetric
nodes, treats algebraically equivalent leaf compositions as
identical, and requires each leaf weight to fall within
$10^{-6}$ of its target.
Numerical match requires held-out RMSE below $10^{-6}$.  The
empirical estimator $\hat R$ averages recovery uniformly over
64 seeds and 4 initialization strategies (256 trials per
$(A, f^\star)$ pair where applicable).  Other sources of
randomness are held fixed at the defaults established by the
sensitivity check in Section~\ref{sec:setup}.

The relation $f^\star \in E(A)$ does not imply $R(A, f^\star) > 0$.
The central hypothesis tested below is that representability is
necessary but not sufficient under the tested gradient-based
training and hardening procedure.  A target may lie in $E(A)$
while still having near-zero empirical recoverability.
The remainder of the paper measures $\hat R$ across architectures,
operators, and target shapes and reports the gap between
representability and recovery.

\begin{table}[t]
\centering
\caption{Architectures studied at depth~3.}
\label{tab:archs}
\footnotesize
\begin{tabular}{@{}llrl@{}}
\toprule
Name & Internal selection & Params & Variable routing \\
\midrule
Eq.\,6 & softmax $\{1,x,f\}$ & 42 & $x$ at all nodes; $y$ at leaves \\
V16    & sigmoid $\{1,f\}$   & 38 & $x,y$ at leaves only \\
Hybrid & V16 + root 4-way    & 44 & $x,y$ at root and leaves \\
\bottomrule
\end{tabular}
\end{table}

%% file: content/methodology.tex
\section{Methodology}
\label{sec:setup}

To test whether the architecture--target interaction is specific to
EML or general to operators with gradient asymmetry, two additional
operators are introduced:
\begin{align}
  \sml(a,b) &= \sinh(a) - \arctan(b), \label{eq:sml} \\
  \rml(a,b) &= \arctan(a) - \sinh(b). \label{eq:rml}
\end{align}
SML shares EML's gradient direction.  The left input receives
unbounded gradients ($\partial\sml/\partial a = \cosh a$, which
grows exponentially) while the right is bounded
($\partial\sml/\partial b = -1/(1+b^2)$).  RML reverses this
asymmetry.  Neither SML nor RML is known to be universal, and target trees are evaluated numerically.

Each node's left and right inputs are weighted combinations of
an architecture-specific set of available terms, typically drawn from
$\{1, x, y, f_{\text{child}}\}$.  Which terms are available at which
nodes is exactly what distinguishes the three architectures
(Table~\ref{tab:archs}).  Weights are controlled by learnable
parameters under a softmax or sigmoid.  Training minimizes mean
squared error using Adam with a learning rate
of 0.01 on a $21 \times 21$ grid over the positive quadrant
$[1,3]^2$ (441~points).  The domain is bounded away from zero
for two reasons.  First, EML's right-input logarithm $\ln b$
requires $b>0$.  Second, the singularity in
$\partial\eml/\partial b = -1/b$ near zero would destabilize
gradient measurements.  Targets with
numerical overflow on a subset of the grid are excluded entirely,
leaving only targets whose function values are finite on all
441~points.  Training proceeds in two phases.  A
search phase runs for 6{,}000 iterations at temperature $\tau = 2.5$, with
soft softmax and sigmoid weights, followed by a hardening phase
of 2{,}000 iterations in which $\tau$ anneals from 2.5 to 0.01 with entropy
and binarity penalties before the one-hot snap.  After hardening
each node selects exactly one input, producing a concrete
expression tree.

Each combination of architecture, operator, tree shape, and leaf
assignment is one cell.  V16 and Hybrid are tested with 256
trials per cell, using 64 seeds over four initialization
strategies.  Eq.\,6 is tested with 64 trials per cell using the
original paper's randn~$\times$~0.1 initialization.  Applying
V16's initialization set to Eq.\,6 would mix architecture and
initialization effects.  V16 and Hybrid initializations bias the
softmax weights either toward the constant~$1$ (\emph{biased}),
uniformly across inputs (\emph{uniform}), toward the variable
inputs (\emph{$xy$-biased}), or onto a single randomly chosen
input (\emph{random-hot}).  All reported rates carry 95\% Clopper--Pearson
confidence intervals (at $n = 256$,
mid-range half-widths $\approx$6\,pp; at $n = 64$,
0\% yields $[0, 5.6]$\%).  A trial counts as exact recovery when
the hardened model satisfies training MSE below $10^{-20}$ on the
441-point grid and held-out RMSE below $10^{-6}$ on 500 points
drawn uniformly from $[0.5,5.0]^2$, which strictly contains the
training grid.  In practice, this tolerance separates exact
symbolic recovery from close numerical fits on the noiseless
grid, and the held-out grid checks both in-range fit and
extrapolation.
Per-trial logs (twelve fields including seed, initialization,
architecture, target, operator, training and held-out RMSE,
hardened expression, routing, structural-match flag, failure-type
label, and wall time) are recorded for every run.

A sensitivity screen varied six protocol factors: optimizer,
hardening schedule, clamp magnitude, held-out grid seed,
training noise $\sigma$, and numerical $\varepsilon$.  The screen
ran 5{,}248 OFAT trials plus a factorial across the twelve
non-balanced chain cells.  The largest cell-dependent effects
came from initialization strategy.  One harmful level appeared
for the optimizer, Adam~$10^{-3}$, and one appeared for the
hardening schedule, step annealing.  The remaining levels
changed recovery by less than $12$ percentage points.  The main experiment
uses Adam~$10^{-2}$, power-law hardening, clamp~$=10^6$, grid
seed~$=12345$, $\sigma=0$, and $\varepsilon=10^{-12}$.\footnote{Source
code: \url{https://github.com/ChakshuGupta13/lab/}.}

%% file: content/results.tex
\section{Results}
\label{sec:results}

\subsection{Architecture Choice Reverses Recovery Across Targets}
\label{sec:results-interaction}

Since Eq.\,6 and V16 use different native initialization
protocols (Section~\ref{sec:setup}), recovery-rate comparisons
are best read as architecture-plus-native-training-protocol
comparisons unless otherwise noted.  The results below separate
three questions: whether architecture changes recovery, whether
the effect persists across operator asymmetry, and whether
held-out validation can reduce architecture-specific failures.
Fig.~\ref{fig:heatmap} shows that recovery cannot be attributed
to just architecture or target.  Recovery is governed by the
specific architecture--target pairing.
Eq.\,6 acts as a specialist with 100\% recovery on two targets, 0\% on
three, and below 2\% on the remaining one.  V16 recovers all four
chain targets at rates between 16\% and 99.6\%, with $x \leftrightarrow
y$ symmetry inherited from its leaf-only variable design.  Hybrid is
an RL specialist, recovering 95--96\% on RL targets and near 0\%
elsewhere.  No single architecture dominates.  Eq.\,6 reaches 100\%
where V16 reaches 17\%, yet V16 reaches 99.6\% where Eq.\,6 reaches
0\%.

In these tested cells, the reversals cannot be explained by
representability.
Eq.\,6's representable set is a strict superset of V16's (every
internal node in Eq.\,6 offers $\{1,x,f_{\text{child}}\} \supset
\{1,f_{\text{child}}\}$, so every V16 tree is also an Eq.\,6
tree), yet Eq.\,6 fails on targets V16 recovers nearly
deterministically.  The failure is therefore not that the
expression is unavailable to the architecture.  The training and
hardening process does not reliably reach the representable
target expression.  Eq.\,6 routes
$x$ through every internal node, so LR targets, whose active
subtree sits on the exponentially amplified left input, receive
strong $x$-gradients early in training, while RL targets place the
active subtree on the attenuated right input and starve $x$ of
signal along the path that must be learned.  Here, starvation
refers to persistently lower gradient norm on the variable path
needed to recover the target.  V16 may avoid this
failure mode because $x$ and $y$ enter only at the leaves through a
shared softmax.  Eq.\,6 nonetheless reaches 100\% on the RR target
with leaves $(y,x)$ despite right-side routing at both levels.
This indicates that leaf assignment interacts with routing
pattern, and that a full account of which combinations succeed
remains open.  Section~\ref{sec:results-gradient}
quantifies the routing imbalance through gradient ratios.

Tracing where each run fails locates the bottleneck.  Across
5{,}376 runs aggregated over initialization strategies, 77\%
reach soft-phase loss below $0.01$, but only 32\% survive the
one-hot snap into a hardened expression with training MSE below
$0.01$, and every one of those then meets the stricter exact-match
tolerances.  The one-hot snap, rather than the final tolerance
threshold, is the main gate.  Failure modes separate by cell.  Eq.\,6/RL under EML
fails during search (soft loss never $<0.01$).  Eq.\,6/RR under
EML reaches a soft fit, but hardening destroys that fit.  EML balanced
cells diverge to $10^{16}$ during search.  SML balanced cells
usually fail during hardening through chain collapse or
snap-induced mismatch.

\subsection{Operator Swap Reverses Shape Preferences}
\label{sec:results-operators}

\begin{figure}[t]
\centering
\includegraphics[width=\columnwidth]{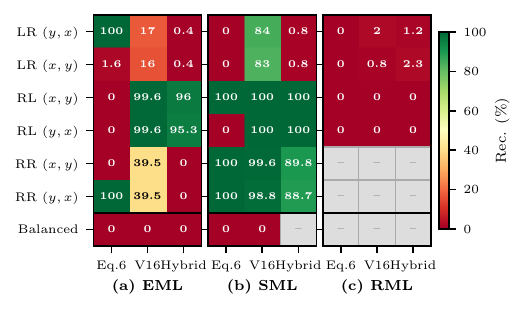}
\Description{Heatmap showing recovery rates for three operators across architectures and tree shapes.}
\caption{Exact recovery after hardening depends on the
architecture--target pairing rather than on architecture or
target alone.  Rows are target shapes and leaf assignments;
columns are architectures; panels are operators (EML, SML, RML).
Gray cells: configurations not tested.}
\label{fig:heatmap}
\end{figure}

The asymmetry-direction hypothesis predicts that operators
sharing EML's left-amplification should preserve some
architecture--shape preferences, while reversing the asymmetry
should change those preferences.  Switching from EML to SML
(Fig.~\ref{fig:heatmap}) changes Eq.\,6's preferred shapes,
most clearly on LR and RL.  The LR rate drops from 100\% to 0\%
and the 100\% targets shift to RL$(xy)$ and both RR variants
(Fisher exact $p < 10^{-15}$).  RR remains leaf-assignment
dependent rather than uniformly easy or difficult.  At least one
RR leaf assignment reaches 100\% for Eq.\,6 under both
operators, but the other scores 0\% under EML
(Fig.~\ref{fig:heatmap}).  Since both operators amplify the
left input, the change on LR and RL is consistent with
differences in right-side attenuation ($-1/(1+b^2)$ for SML
versus $-1/b$ for EML).  The
same per-architecture pattern holds across the swap.  Eq.\,6
remains a specialist, V16 recovers 83--100\% on all chains, and
Hybrid remains an RL specialist.

Under EML the right gradient $-1/b$ on $[1,3]$ is moderately
attenuating, enough for V16's symmetric routing to recover RL
targets but not for Eq.\,6's $x$-dominant path.  Under SML the right
gradient $-1/(1+b^2)$ saturates more severely while the left
amplification weakens from $e^a$ to $\cosh a$; the net effect
appears to shift which shapes give Eq.\,6 a favorable balance of gradient
signal between $x$ and $y$ leaves (quantified in
Section~\ref{sec:results-gradient}).  V16's robustness to the
swap is consistent with its leaf-only routing, which does not
commit either variable to a fixed depth.

V16 dominates under SML, recovering all chain targets at 83\% or
above.  Replacing EML with the right-amplified RML collapses
recovery across the board (Eq.\,6: 0\% on every target; V16: 2.0\%
best, $5/256$, 95\% CI $[0.6, 4.5]\%$; Hybrid: RL preference
gone), supporting the interpretation that gradient direction,
rather than operator identity, contributes to the routing
bias.

\subsection{Balanced Trees Resist Recovery Across All Tested Configurations}
\label{sec:results-balanced}

Balanced topologies fail in all tested configurations
($0/3{,}776$; Clopper--Pearson 95\% upper bound $0.10\%$), even
when the same architecture--operator pair recovers chain targets
at high rates.  The failure persists across six training
configurations varying learning rate, iteration budget, and
temperature ($0/192$).  Failure modes differ by operator.  EML
balanced cells diverge during search (loss reaches $10^{16}$),
whereas SML balanced cells often reach a soft fit but collapse
during hardening (44\% chain collapse, 55\% hardening proper).
This suggests that balanced failure is a training-and-hardening
pathology rather than a simple absence of representability.  Branch-wise
gradient measurements in Section~\ref{sec:results-gradient} show
balanced cells with $\rho_{LR}\approx 2$--$3$, several-fold
weaker than the $10$--$17\times$ asymmetry observed on recovered
chains.  This suggests that neither branch receives the
concentrated signal observed in successful chain recoveries.

\subsection{Gradient Asymmetry Correlates with Recovery}
\label{sec:results-gradient}

Two gradient ratios summarize the asymmetry.  The variable-wise
ratio
\begin{equation}
  \rho_{xy}(t) \;=\; \frac{\|\nabla_{\theta_x} \mathcal{L}_t\|}{\|\nabla_{\theta_y} \mathcal{L}_t\|}
  \label{eq:rho-xy}
\end{equation}
compares signal strength on the two variable leaves, and the
branch-wise ratio
$\rho_{LR}(t) = \|\nabla_{\theta_{\text{left}}} \mathcal{L}_t\| \,/\,
\|\nabla_{\theta_{\text{right}}} \mathcal{L}_t\|$
compares signal on the two children of an internal node.  Both are
measured at iteration~1000 across 10 seeds and reported jointly in
Table~\ref{tab:gradients}.  Gradient measurements are reported for the
two most extreme Eq.\,6 targets per operator (100\% and 0\%); the
intermediate targets (T4, T8) are omitted because their mixed
recovery depending on leaf assignment makes the association
ambiguous.  For Eq.\,6 under EML, $\rho_{xy}$ tracks the two
extreme outcomes: $\rho_{xy} < 1$ appears on the 100\% target,
while $\rho_{xy} > 1$ appears on the 0\% target.  Under SML the
correspondence is more nuanced (EML LR: $0.67 \to 100\%$; SML LR:
$1.78 \to 0\%$; SML RL: $1.19 \to 100\%$): the relevant variable
depends on the operator--shape combination, not a universal
threshold.  V16 tolerates a $2.3\times$ imbalance on a 99.6\%
target, suggesting its symmetric leaf-routing provides alternative
pathways that compensate for ratio skew.

\begin{table}[t]
\centering
\caption{Gradient diagnostics at iteration~1000 (10-seed mean
$\pm$ s.e.).  Leaf $\rho_{xy}$ is the variable-wise ratio at
the leaves; branch $\rho_{LR}$ is the left/right gradient-norm
ratio at the root (V16).}
\label{tab:gradients}
\footnotesize
\begin{tabular}{@{}lllrc@{}}
\toprule
Diagnostic        & Arch.\,/Op.\ & Target     & Ratio          & Rec.\ \\
\midrule
Leaf $\rho_{xy}$  & Eq.\,6 EML  & LR\,(yx)  & $0.67\pm0.03$  & 100\% \\
Leaf $\rho_{xy}$  & Eq.\,6 EML  & RL\,(xy)  & $1.58\pm0.04$  & 0\% \\
Leaf $\rho_{xy}$  & Eq.\,6 SML  & LR\,(yx)  & $1.78\pm0.12$  & 0\% \\
Leaf $\rho_{xy}$  & Eq.\,6 SML  & RL\,(xy)  & $1.19\pm0.11$  & 100\% \\
Leaf $\rho_{xy}$  & V16 EML     & RL\,(xy)  & $2.31\pm0.24$  & 99.6\% \\
\midrule
Branch $\rho_{LR}$ & V16 EML    & LR\,(yx)  & $1.20\pm0.06$  & 16.8\% \\
Branch $\rho_{LR}$ & V16 EML    & RL\,(xy)  & $0.06\pm0.01$  & 99.6\% \\
Branch $\rho_{LR}$ & V16 SML    & LR\,(yx)  & $10.0\pm1.2$   & 84.4\% \\
Branch $\rho_{LR}$ & V16 SML    & RL\,(xy)  & $0.07\pm0.00$  & 100\% \\
Branch $\rho_{LR}$ & V16 EML    & Bal\,(xy) & $3.23\pm0.29$  & 0\% \\
Branch $\rho_{LR}$ & V16 SML    & Bal\,(xy) & $2.62\pm0.70$  & 0\% \\
\bottomrule
\end{tabular}
\end{table}

\begin{figure}[t]
\centering
\includegraphics[width=\columnwidth]{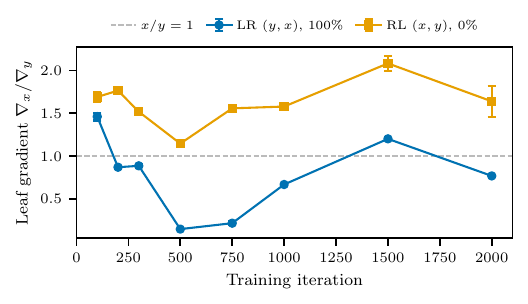}
\Description{Leaf x/y gradient ratio trajectory for Eq.6 on a recovered vs an unrecovered target during search-phase training.}
\caption{Leaf-gradient ratio $\rho_{xy}(t) =
\|\nabla_{\theta_x} \mathcal{L}_t\| /
\|\nabla_{\theta_y} \mathcal{L}_t\|$ during Eq.\,6 search under
EML for a recovered LR target ($100\%$ recovery) and an
unrecovered RL target ($0\%$); 10-seed mean $\pm$ s.e.}
\label{fig:gradient}
\end{figure}

The gradient trajectory (Fig.~\ref{fig:gradient}) shows the
pattern more clearly.  On the 100\% EML target $\rho_{xy}$ briefly
reverses, falling from $1.45$ at iteration~100 to $0.14$ at
iteration~500 before settling at $0.67$ by iteration~1000, and this
$y$-dominated window correlates with convergence.  On the 0\% target
$\rho_{xy}$ stays above $1.14$ throughout.  The trajectory provides
a gradient-level correlate of the architecture--target interaction.
Whether this gradient-level correlate is causal or consequential
is left to Section~\ref{sec:limitations}.

Across the measured configurations, recovery aligns more with the
distribution of gradient signal during training than with any
fixed threshold on $\rho_{xy}$.  V16 recovers when $\rho_{xy}$ sits
near unity, indicating balanced access to both leaves.  Eq.\,6 fails
on RL/EML where $\rho_{xy}$ stays at $1.58$, indicating $y$
starvation.  The same architecture recovers on LR/EML where
$\rho_{xy}$ briefly dips to $0.14$, well below unity, opening a
window in which $y$ can compete.  An operator swap that flips the gradient asymmetry
flips $\rho_{xy}$ and the recovered shape together.  Balanced trees
fit the same picture.  Their branch-wise ratio sits at $\rho_{LR}
\approx 2$--$3$ (Table~\ref{tab:gradients}), several-fold weaker than
the $10$--$17\times$ asymmetry seen on recovered chains, so neither
branch receives the concentrated signal that chain recovery needs.
Both SML balanced variants stay left-biased regardless of $x
\leftrightarrow y$ swap, pointing to a structural rather than
target-driven origin in the operator's left amplification.  These
patterns remain a measurable correlate, not a proven causal
mechanism (Section~\ref{sec:limitations}).

\begin{figure}[t]
\centering
\includegraphics[width=\columnwidth]{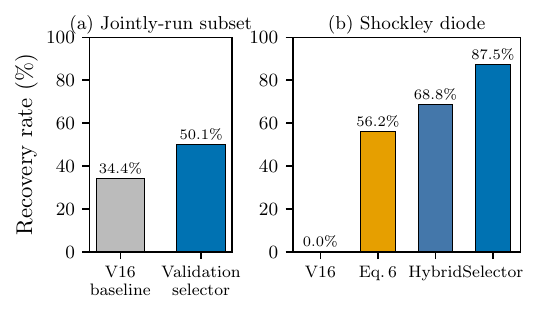}
\Description{Validation selector recovery compared to individual architectures on jointly-run synthetic targets and the Shockley diode.}
\caption{Held-out validation over a small architecture set
reduces architecture-specific recovery failures.  For each
target, the candidate with the lowest held-out RMSE is selected.
On the jointly-run subset, selection improves recovery over the
V16-only baseline; on a Shockley diode target, selection
combines complementary successes from Eq.\,6 and Hybrid.}
\label{fig:selector}
\end{figure}

\subsection{Validation-Selected Architecture and a Practical Target}
\label{sec:results-selection}

The reversals in Section~\ref{sec:results-interaction} suggest
a simple proof-of-concept mitigation.  Train a small candidate
set of architectures on the target, harden each candidate, and
select the expression with the lowest held-out RMSE.  The selector is evaluated on the subset of configurations where
V16 and at least one other architecture were both run: EML
LR\,$(y,x)$ (Eq.\,6 + V16), RML LR\,$(y,x)$ (Eq.\,6 + V16), and
SML LR\,$(y,x)$ (V16 + Hybrid).  Pooling
the $256$ common (seed, initialization) trials per
configuration, V16, the only single architecture present
on all three, recovers $264/768 = 34.4\%$.  The validation
selector recovers $385/768$ trials, or $50.1\%$
(Fig.~\ref{fig:selector}a).  Since this subset contains only
three configurations, the result should be read as evidence that
validation-based architecture selection is promising, not as a
complete benchmark.  On this subset, the validation selector
matches the per-target oracle because exact and non-recovery
runs separate by roughly five orders of magnitude in held-out
RMSE.  Re-running V16 with
Eq.\,6's initialization strategy on SML LR\,$(y,x)$ collapses
its rate from $84.4\%$ to $0/64$, so a deployable selector must
hold the per-architecture initialization set fixed.

As a representativeness check, we evaluate the Shockley diode
equation $I = I_0\,(\exp(qV/kT) - 1)$~\citep{Shockley1949pn},
a classic nonlinear electrical-engineering model with
signal-processing relevance.  The normalized Shockley diode
form, $\mathrm{eml}(x,e)$, is depth-1 in EML, so the diode
equation provides a useful target outside the synthetic target
family.
Under the same depth-3 protocol with $x \in [0, 5]$, Eq.\,6
recovers the target on $18/32$ seeds, Hybrid on $22/32$ seeds,
and V16 on $0/32$ seeds.  The selector chooses Eq.\,6 or Hybrid
in every recovered run and recovers $28/32$ seeds by covering
complementary successes across architectures
(Fig.~\ref{fig:selector}b).  V16's failure is representational
rather than optimizational.  V16 takes the variable~$x$ only at
leaves, so the root's left input must be either a constant or
the output of a depth-$\le 2$ V16 subtree, and exhaustive
enumeration of those outputs over $\{1, x, y\}$ contains no
expression equal to $x$.  In this experiment, a single fixed
V16 choice would yield zero recoveries.

%% file: content/discussion.tex
\section{Discussion}
\label{sec:discussion}

These results should be read as a controlled study of
recoverability in depth-limited differentiable symbolic
regression, not as a benchmark ranking symbolic regression
methods broadly.  For closed-form signal-model discovery,
recoverability depends not only on the operator grammar but also
on the architecture through which signal variables are routed
during training.  Varying the architecture's variable-routing
pattern while holding operator, tree shape, and leaf
assignment fixed produces 0\%--100\% recovery reversals across
three operators, three architectures, and twenty-four target
configurations.  What each architecture can represent does not
fully explain the pattern.  Eq.\,6's representable set is a strict superset
of V16's, yet Eq.\,6 fails on targets V16 recovers at 99.6\%.
Balanced topologies fail to recover under any combination tested
(0/3{,}776), and reversing the operator's input asymmetry collapses
all rates to 2\% or below.  These results provide evidence for architecture-induced
recoverability bias as a measurable property of differentiable
symbolic regression on the configurations tested.  Recovery
differences are associated with variable routing and
optimization-landscape geometry, a pattern most directly
observable when the variable-routing architecture is isolated, a
condition absent from benchmarks where methods vary
simultaneously in grammar, search, and
architecture~\citep{LaCava2021srbench}.  This parallels findings
in differentiable architecture search~\citep{Liu2019darts},
where smooth relaxations can yield unstable discrete
selections~\citep{Zela2020robustdarts}.

Cross-operator results suggest that recovery moves with variable
routing, tree shape, leaf assignment, and gradient direction
together rather than with any one factor in isolation.  Reversing the
operator's input asymmetry with RML collapses Eq.\,6 and Hybrid
to $0\%$ on every tested chain rather than mirroring their EML
preferences.  This is consistent with left-amplifying asymmetry
being important for Eq.\,6 recovery under this protocol.

\subsection{Practical Implications}
\label{sec:practical}

The validation-selected ensemble in
Section~\ref{sec:results-selection} suggests a simple mitigation
strategy.  In nonlinear system identification and interpretable
signal modeling, the goal is often recovery of a reusable
closed form rather than just low prediction error, so
architecture choice can determine whether a physically
meaningful expression is discoverable at all.  A practitioner
can train a small architecture set, harden each candidate, and
select by held-out RMSE.  This is a proof of concept rather than
a complete deployment rule, since the selector was evaluated on
a limited subset and on noiseless targets where recovery and
non-recovery separate clearly.  Two qualifications matter.
First, the lift comes from two sources: cell-level selection
($34.4\%$ for V16 to $50.1\%$ on the three jointly-run
configurations) and coverage, where adding architectures avoids
commitment to a candidate whose representable set excludes the
target, as the Shockley diode case isolates.  Second, the
selector matches the per-target oracle only because recovery
and non-recovery runs separate by five orders of magnitude in
held-out RMSE.  On noisier data or tighter training budgets the
gap may shrink and the selector may pick a non-recovering run
that happens to fit slightly better.

\subsection{Limitations}
\label{sec:limitations}

The architecture--target interaction is established only at depth~3
under the subtraction form $f(a)-g(b)$.  The results should
therefore not be read as a claim about all differentiable SR
architectures or all symbolic grammars.  Multiplicative and
compositional operators are untested, and a limited depth-4 V16
sweep suggests the depth-3 generalist profile may not persist
(RL: $99.6\%\to 0.8\%$), so a full depth-4 matrix is needed
before the depth-3 ordering can be treated as general.  The
narrower output range of SML on $[1,3]^2$ may partly explain
V16's higher SML rates, but the shape-specific reversals
(Eq.\,6 under SML: LR~$0\%$, RL~$100\%$) are unlikely to be
explained by only output conditioning.  The paper does not
claim EML or SML is a strong practical grammar for
signal-processing formula discovery; instead, these grammars
are used as controlled settings in which representability and
recoverability can be separated.

The gradient ratio $\rho_{xy}$ is a diagnostic correlate
(Section~\ref{sec:results-gradient}), not a causal account.  The
sensitivity screen (Section~\ref{sec:setup}) indicates that the
observed architecture--target reversals are not artifacts of
optimizer, clamp, held-out seed, training noise, or numerical
$\varepsilon$ under the tested ranges, although initialization
remains cell-dependent.

The signal-model audit shows where the controlled recovery
phenomenon intersects with common formula classes used in
electrical, acoustic, and communication modeling.
Table~\ref{tab:audit} audits several canonical signal-processing
nonlinearities at depth~3.  Some are directly representable and
tested as recovery targets; others expose grammar barriers
involving addition, reciprocal structure, signed exponents, or
constant construction.

Many canonical two-variable signal-processing,
electrical-engineering, and physics formulas fall
into the second group.  Addition in EML requires a much deeper
tree~\citep{Odrzywolek2026eml}, and the current
softmax-weighted leaves cannot represent negation.  Constant
constructibility also matters.  With the current leaf alphabet
$\{1,x,y\}$, zero cannot be synthesized as a subtree on the
training domain.  As a result, log-only responses of the form
$\eml(0,\cdot)$ are structurally unreachable, while exponential
responses such as the normalized Shockley diode remain
reachable.

\begin{table}[t]
\centering
\caption{Signal-model audit at depth~3.  R: representable;
A: blocked (missing primitive in parentheses).}
\label{tab:audit}
\footnotesize
\begin{tabular}{@{}lll@{}}
\toprule
Formula class (example) & Application context & Status \\
\midrule
Exp.\ device law (Shockley) & electronics; sensors & R \\
Log compression ($1-\ln x$) & dB; log spectra      & A (zero leaf) \\
Friis / path loss           & RF; comms            & A (sum) \\
Shannon capacity            & comms; info rate     & A (sum) \\
Sabine / Hill               & acoustics; sensory   & A (reciprocal) \\
\bottomrule
\end{tabular}
\end{table}

%% file: content/conclusion.tex
\section{Conclusion}
\label{sec:conclusion}

In depth-limited differentiable symbolic regression,
architecture is not merely an implementation detail.  Even when
operator and grammar are controlled, the variable-routing
pattern can strongly affect whether a representable expression
is recovered after training and hardening, with recovery moving
between $0\%$ and $100\%$ on the same target in the tested
configurations.  The practical implication is direct.
Architecture should be reported, stress-tested, and selected by
held-out validation rather than fixed a priori.  The
next step is to test whether this recoverability bias persists
in deeper trees and in grammars with addition, multiplication,
and composition as primitives.